# Adversarial Security Attacks and Perturbations on Machine Learning and Deep Learning Methods

Arif Siddiqi, Ph.D.
Adjunct Faculty
a.siddiqi@snhu.edu | emailarif@aol.com
Southern New Hampshire University
College of Online and Continuing Education (COCE)
2500 N River Rd,
Hooksett, NH 03106 USA

# Adversarial Security Attacks and Perturbations on Machine Learning and Deep Learning Methods

## ABSTRACT

The ever-growing big data and emerging artificial intelligence (AI) demand the use of machine learning (ML) and deep learning (DL) methods. Cybersecurity also benefits from ML and DL methods for various types of applications. These methods however are susceptible to security attacks. The adversaries can exploit the training and testing data of the learning models or can explore the workings of those models for launching advanced future attacks. The topic of adversarial security attacks and perturbations within the ML and DL domains is a recent exploration and a great interest is expressed by the security researchers and practitioners. The literature covers different adversarial security attacks and perturbations on ML and DL methods and those have their own presentation styles and merits. A need to review and consolidate knowledge that is comprehending of this increasingly focused and growing topic of research; however, is the current demand of the research communities. In this review paper, we specifically aim to target new researchers in the cybersecurity domain who may seek to acquire some basic knowledge on the machine learning and deep learning models and algorithms, as well as some of the relevant adversarial security attacks and perturbations.

**Keywords:** Machine learning algorithm; adversarial security attack; adversarial perturbation; adversarial example; vulnerabilities; exploitations.

**Highlights:**

- Machine learning methods are vulnerable to adversarial security attacks
- Adversarial examples, poisoning, and exploratory attacks are most common
- SVM, NN, and DNN are popular ML/DL methods to study adversarial security
- MNIST, CIFAR, and ImageNet are popular choices of datasets
- Research in Adversarial Security Attacks and Perturbations is fast growing

## 1. Introduction

### *1.1 Introduction*

The domain of artificial intelligence (AI) and its subdomains of machine learning (ML) and deep learning (DL) have made a tremendous impact on our lives in recent years. A wide variety of products and applications in nearly every sector exist where the technologies of artificial intelligence are the driving force. Amazon Alexa, Apple SIRI, and Tesla smart car are some of the few generic examples of these technologies. The growth and development of these technologies themselves are attributed to the advancement in information technology (IT) sector and the big data as a result of expansion of Internet. Artificial intelligence (AI) which is the science of making a machine intelligent without explicitly programmed is one emerging market and hot research domain. The technologies of AI can process big data but also harness a huge computing power of the machine. Behind all this the models and algorithms of machine learning and deep learning working on varieties of data types and sizes make intelligent decisions and allow the machine becoming human like a reality.

The ML and DL have been used in the applications of pattern recognition, image recognition, recommendation and filtering, network intrusion detection, malware detection, spam detection, data mining, cryptosystems, physically unclonable function (PUF), self-driving cars, unmanned autonomous systems, electric grid, and the areas of medicine and healthcare, homeland security, cybersecurity, marketing, sales, etc. (Bulò et al., 2017; Che et al., 2018; Duddu, 2018; Liu et al., 2018; Mispan et al., 2018; Padilla, Meyer-Baese, & Foo, 2018; P, Kumar, & T, 2016; Sahingoz et al., 2018; Sahoo et al., 2015; Sethi & Kantardzic, 2018b; Su, Zwolinski, & Halak). Some of these applications pertain to critical infrastructures that have national interests and strategic importance, whereas others have their influence or mark globally (Duddu, 2018). As with any technology, the design and development of ML and DL models or algorithms is a continuous process and requires input from the end users for improvements (Liu et al., 2018). The data that those models and algorithms are fed sometimes distributed over the computers and networks or of completely unknown character and behavior. The security, therefore, is a concern or become important in meeting the requirements or validating the accuracy of the models or results. Every process of the ML and DL is vulnerable to attack from the training to testing phase and to the design of the model or algorithm itself.

The models of machine learning traditionally assumed the environment was benign, because of the limited computing platforms and the simple codes execution of those models on the training and test data sets (Goodfellow, McDaniel, and Papernot, 2018); however, in the adversarial domain and with the advancement of technologies, that assumption does not hold valid and the distributions of both the training and test data sets may significantly differ (Cárdenas, 2012; Sethi & Kantardzic, 2018a,b). Adversaries armed with the abundant of open source and commercial tools on the Internet and with an evil intent can do a variety of damages to the ML and DL models or their claimed results, without the designers or operators of those noticing any wrongdoings for some time (Liu et al., 2018). For example, the control of the autonomous cars; bypassing the spam filters, fake recommendations, exploitation of facial and voice recognition systems, etc. are all the possibilities (as cited in Liu et al., 2018).

Some scholars in the early-to-mid 2000 brought the much-needed attention and contributed on the issue of security in ML and DL models and algorithms. A few noted seminal work in this area includes: Dalvi's (2004) concept of adversarial classification and evasion of linear classifiers; Lowd and Meek's (2005) concept of adversarial learning and adversary-aware classifiers; Barreno et al.'s (2006) views on ML problems and taxonomies, etc. (as cited in Biggio & Roli, 2018; Liu et al, 2018; Li et al., 2018). Since then the work on adversarial security attacks and perturbations is on-going and is one of the research topics of interest among the scholars of cybersecurity and artificial intelligence (Corona, Biggio, & Maiorca, 2016).

There have been several publications such as Akhtar and Mian (2018), Barreno et al. (2010), Bae et al. (2019), Chakraborty et al. (2018), Duddu (2018), Gardiner and Nagaraja (2016), Li et al. (2018), Liu et al. (2018), Kumar & Mehta (2018), Yuan et al. (2018), Zhang et al. (2019), etc. that cover the literature reviews of adversarial security attacks and perturbations. They all bring useful information on this topic that is unique and meritorious. Some provide in-depth coverage on the theoretical aspects of various adversarial security attacks types, whereas others focus more on the perturbations from the perspectives of computer-vision (CV) researchers. Furthermore, some efforts have also been towards the quantification and design of the classifiers in the adversarial domains, based on the feature selection or engineering (Eykholt, K., & Prakash; Katzir & Elovici, 2018). The research in the domains of artificial intelligence and its subdomains of ML and DL is increasing at an unprecedent level and most definitely the coverage of the security aspects will go together (Carlini et al., 2019; Joseph et al., 2012; Newcombe, 2018; Rieck, 2016; Thornton et al., 2015). We anticipate an increased number of research publications in these topics and that includes reviews covering different perspectives and coverage over the years. In this review paper, we aim to target new entrants of cyber security in the research area who may seek to acquire some basic knowledge on the machine learning and deep learning models and algorithms, as well as some of the relevant adversarial security attacks and perturbations. We avoided providing extensive detail on each security attack or perturbation type, comprising of mathematical equations and theoretical aspects; however, the readers are expected and encouraged to consult the references listed for guidance on this topic.

The presentation of this paper is as follows. Section 2 provides a brief overview on the background and types of machine learning and deep learning methods, as well as a list of those with small descriptions. Section 3 provides a brief overview containing the classification schemes, taxonomies, definitions, and terms of adversarial security attacks and perturbations on ML and DL systems, as well as a list of those with small descriptions. Section 4 provides a comparison analysis of selective security attacks, ML/DL methods, datasets, and their relevant research studies. Section 5 provides discussion and conclusion based on the findings and some directions for future research. Appendix - in the last section - provides a table with a comprehensive list of security attacks and perturbations.

## 2. Machine learning (ML) and deep-learning (DL) Methods

### *2.1 Brief Overview*

Machine learning (ML) and deep-learning (DL) are a subset of bigger technology group called artificial intelligence (AI). The history of artificial intelligence together with the machine learning and deep neural network (DNN) (a precursor to deep learning) goes back to the 1940s and since then the work is ongoing in the field with some big marks in the 1980s (Alom et al., 2018). Now with the advancement in computer and related technologies, a growing and regenerated interest and development can be seen again in recent years. Deep learning (DL) is the latest creation with works such as the Hinton lab's training for DNNs in 2006 and AlexNet by Alex Krizhevesky in 2012 (as cited in Alom et al. (2018)). Artificial intelligence (AI) is the science of making a machine intelligent that closely exhibits the character of human in terms of thinking and abilities without explicitly programmed. Machine learning (also known as shallow learning) and deep learning with their learning methods that consists of algorithms and models and the use and input of different data types and sizes allow to handle and automate AI-related tasks such as pattern recognition, decision-making, observations of a dynamic environment without explicit instructions (Ateniese et al., 2015; Xue & Chua, 2018). ML and DL algorithms, models, designs and architectures differ in some respects. For example, ML algorithms lack automatic feature engineering and have low detection rate, inability of detecting small perturbations in the sample, or handling large datasets, however, DL algorithms or models can overcome some of these limitations or challenges (Xin et al., 2018). DL uses the representation learning for a feature selection and can handle large and multi-layered datasets (Apruzzese et al., 2018; Diro & Chilamkurti, 2018). The superiority of algorithms and training data sets affect the effectiveness of trained machines or the performance of the models and therefore, are sometimes guarded as a trade secret (at least the data sets) (Ateniese et al., 2015).

ML and DL methods, networks, models, or systems - depending on the algorithms, data type, and tasks - can be categorized, according to Alom et al. (2018) and Xin et al. (2018), as follows:

- Supervised. The input data is labeled for training and the learned procedure is used on the test data.
- Semi-supervised. The combined method of supervised and unsupervised that reduces the label efforts or uses the partial labeled data, providing high accuracy of the algorithms in large data.
- Unsupervised. The input data is unlabeled, and the key features of data are described or summarized.
- Reinforcement learning. The method that requires exploration and applies to unknown environments; sometimes also called semi-supervised.

Also, ML and DL methods are based on a single classifier, hybrid classifier (cascading two or more classifiers), and ensemble (a group of weak algorithms or weak learners; use of data resampling techniques such as boosting and bagging and the strategy of majority vote) classifier (Tsai et al., 2009).

### *2.2 Machine Learning Methods – Supervised*

#### *2.2.1 Naive Bayes (NB)*

A probabilistic classifier or model that assumes the input features are independent from each other (Apruzzese et al., 2018). Also, NB estimates the parameters based on the maximum likelihood principle. The classifier is simple and scalable and can produce fast results on a small training data (Guerrero-Higueras, DeCastro-García, & Matellán, 2018). The classifier uses conditional probability formula to answer the probability of a certain event (Nguyen, 2018; Tsai et al., 2009). NB is typically represented by a directed acyclic graph (DAG) having nodes and links (Tsai et al., 2009).

#### *2.2.2 Logistic Regression (LR)*

A categorical classifier that is based on a discriminative model and assumes the input features are independent to each other (Apruzzese et al., 2018). The target classified value in this method is expected to be a linear combination of the input variables (Guerrero-Higueras et al., 2018). The model uses a logistic function and coordinate descent (CD) algorithm (the default in Scikit-Learn software) to classify input variables (Guerrero-Higueras et al., 2018).

#### *2.2.3 Support Vector Machine (SVM)*

A non-probabilistic classifier that was proposed by Vapnik in 1998 (Apruzzese et al., 2018; Tsai et al., 2009). It uses hyperplanes and kernels such as linear, polynomial, and Gaussian Radial Basis Function in high dimensional space to maximize distance between categories of samples and can be used for the tasks of classification, regression, and outlier detection (Chakraborty et al., 2018; Nguyen, 2018). The classifier performs poorly in multi-class classifications and this limited scalability leads to long processing times (Apruzzese et al., 2018); however, the classifier is robust for binary classification of training vectors belonging to two different classes and outlier detection (Tsai et al., 2009). The classifier also uses a penalty factor allowing the tradeoff between the misclassified samples and the width of a decision boundary (Tsai et al., 2009).

*2.2.4 Decision Tree (DT)*

A non-parametric classifier that uses decisions or tree-chart like structure to classify a sample (Tsai et al., 2009; Nguyen, 2018). The classification starts at the root node and finish at the leaf node representing a classification category (Tsai et al., 2009; Gu Guerrero-Higueras et al., 2018). Several well-known implementations of this classifier with different algorithms exists such as CART, ID3, and C4.5. CART can be used as classification or regression tree depending on the categorical or numerical variable (Guerrero-Higueras et al., 2018).

*2.2.5 Random Forest (RF)*

A classier that uses multiple decision trees and provides the unified final response based on those trees (Apruzzese et al., 2018). The classifier uses the technique of the decision tree and an additional vector through boostrap re-sampling (Guerrero-Higueras et al., 2018). The classifier is appropriate for multi-class classification and large datasets; however, it is prone to error such as overfitting (Apruzzese et al., 2018).

*2.2.6 Hidden Markov Model (HMM)*

A model that allows the treatment of a system as a set of states that are hidden and produces outputs with different probabilities (Apruzzese et al., 2018). The model allows to determine the inputs based on the observable outputs with applications in temporal behavior, reinforcement learning, the likelihood of a sequence of events. The model can be worked with both labeled and unlabeled datasets (Apruzzese et al., 2018).

*2.2.7 K-Nearest Neighbor (KNN)*

A non-parametric and one of the most simple and traditional classifiers that computes the approximate distances between different points in the input vectors (Guerrero-Higueras et al., 2018; Tsai et al., 2009). KNN is an instance-based learning that uses input vector and classify new instances without a formal training stage (Tsai et al., 2009). The classifier is suitable for multi-class problems but with some computational costs as it compares all the training sample (Apruzzese et al., 2018). K is an important parameter in the classifier that influences the classification time and accuracy of prediction (Tsai et al., 2009). KNN is the foundational for unsupervised learning; however, the classification of discrete labels in supervised learning has also been used (Guerrero-Higueras et al., 2018). The ML software such as Scikit-Learn implements K-Neighbors classifier based on the nearest neighbors where users can manipulate the K-integer value (5 by default) (Guerrero-Higueras et al., 2018).

*2.2.8 Neural Network (NN) or Shallow Neural Network (SNN)*

The algorithm or network is based on the neural networks which is inspired by the biological neurons or the collection of perceptron (Apruzzese et al., 2018; Chakraborty et al., 2018). The network uses weights via back-propagation algorithm and an activation function to determine the output (Chakraborty et al., 2018). The network is organized in layers of sensory nodes (input) and hidden layers (output) (Guerrero-Higueras et al., 2018; Tsai et al., 2009). The network is used in the environment where the relationship between the input and output is not clear or for recognizing patterns (Tsai et al., 2009; Guerrero-Higueras et al., 2018). Some examples include ML supervised learning (e.g., Multi-layer Perceptron (MLP) and Artificial Neural Network (ANN)), DL supervised learning (e.g., Convoluted Neural Network (CNN) and Deep Neural Network (DNN)) and unsupervised learning (Self-Organizing Maps (SOM)) (Chakraborty et al., 2018; Tsai et al., 2009; Guerrero-Higueras et al., 2018).

*2.3. Machine Learning Methods – Unsupervised*

*2.3.1 Clustering*

Clustering models or algorithms find data points with similar characteristics or patterns in unlabeled multidimensional data and do not require any explicit description of the classes or categories (Apruzzese et al., 2018; Nguyen, 2018). Examples include: Apriori algorithm, K-means, K-nearest neighbors, probabilistic learning - Expectation Maximization (EM) and Gaussian Mixture Model (GMM); dimensionality reduction - Principal Component Analysis (PCA) and Singular Value Decomposition (SVD); density-based - Density-based Spatial Clustering of Applications with Noise (DBSCAN) and DENsity-based CLUstEring (DENCLUE), etc.

*2.3.2 Association Rules*

The rules identify and predict unknown patterns in data (Apruzzese et al., 2018). Sometimes the excessive output of those rules and the validities require human inspection (Apruzzese et al., 2018). Some examples include Apriori, FP-Growth, ECLAT, etc.

*2.4 Deep Learning Methods – Supervised*

*2.4.1 Deep Neural Network (DNN)*

All deep learning algorithms belong to the Neural Networks (NN) or Deep Neural Networks (DNN) and are based on the biological neurons, like the neural network of ML (Apruzzese et al., 2018). The network consists of many layers that are capable of feature extraction or representation (autonomous) learning (Apruzzese et al., 2018).

*2.4.2 Convolutional Neural Network (CNN)*

CNN is a variant of DNN that consists of convolutional or sub-sampling layers and fully connected layers sharing the weights and reducing the number of parameters (Apruzzese et al., 2018; Chakraborty et al., 2018). CNN uses a feature map that is further reduced in dimensionality by pooling or sub-sampling, while retaining important information (Chakraborty et al., 2018). CNN uses feature extraction and can handle tasks related to data and images recognition (Chakraborty et al., 2018).

### *2.4.3 Feed-forward Neural Network (FNN)*

FNN is a variant of DNN where neurons are interconnected between layers and information is forward directional instead of cyclic as in RNN (Apruzzese et al., 2018). A network that can be used for classification without any assumption on the input data but that comes with a computational cost (Apruzzese et al., 2018)

### *2.3.4 Recurrent Neural Network (RNN)*

A variant of DNN that make use of a memory or sequential information. The inputs and outputs are not independent; rather the output is dependent on the previous calculation in the sequence. They are sequence generators with high computational cost and can handle tasks related to natural language processing (NLP), text and image processing (Apruzzese et al., 2018). Some examples include Long-Short Term Memory (LSTM), Gated Recurrent Units (GRU), etc.

## *2.5 Deep Learning Methods – Semi-supervised*

### *2.5.1 Semi-supervised Learning*

Semi-supervised learning (sometimes called Deep Reinforcement Learning (DRL)) is based on the partially labeled datasets and the learning problem in an unknown environment (Alom et al., 2018). A combination of DL such as DRL and Generative Adversarial Networks (GAN), and LSTM and GRU of RNN has been used as a semi-supervised learning (Alom et al., 2018). The learning uses cases can be seen for the decision making in Sciences and Economics, the reward strategy in Neurosciences, and the behavior knowledge of robots, etc. (Alom, et al., 2018). The knowledge of some essential functions and concepts such as Q-Learning, Deep Q-Networks (DQN), Policy Gradient, Transfer Learning are desired for working with DRL.

## *2.6 Deep Learning Methods – Unsupervised*

### *2.6.1 Deep Belief Network (DBN)*

The network uses a Restricted Boltzmann Machine (RBM) which is a two-layer neural network (Apruzzese et al., 2018; Alom et al., 2018). RBM is an undirected generative model that uses the energy function and the hidden layer to explain the distribution of variables of interest (Alom et al., 2018). DBN is good in feature extraction but require training phase and unlabeled datasets (Apruzzese et al., 2018).

*2.6.2 Auto Encoder (AE)*

A class of neural networks that uses the encoder and decoder phases. The input data is mapped to feature representation for task related to dimensionality reduction, compression, fusion, etc. (Alom et al., 2018). There are various auto encoder types such as Stacked Autoencoders (SAE), Sparse AE, Split-Brain AE, etc. (Alom et al., 2018). SAE can work on pre-training tasks, small datasets with high accuracy in results (Apruzzese et al., 2018).

*2.6.3 Generative Adversarial network (GAN)*

GAN - developed by Goodfellow in 2014 - is a zero-sum game between two neural networks or between the two players, Discriminator (output) and Generator (input) (Alom et al., 2018). The discriminative network is responsible for distinguishing the original dataset and the one that is produced by GAN, whereas the generative deep network, taking the sample from the discriminative network and some noise added, is responsible for producing adversarial examples that are very close to the original set (Chakraborty et al., 2018). GAN has been used in several domains or has applications such as realistic images of objects, game development, motion development, etc. (Alom et al., 2018). GAN has been used as a semi-supervised or unsupervised learning and its many improved versions exist such as Deep Convolutional GAN (DCGAN), Coupled Generative Adversarial Network (CoGAN), Bidirectional Generative Adversarial Networks (BiGANs), etc. (Alom et al., 2018). Some other examples of GAN include MalGAN, APE-GAN, DCGANs by Radford et al., GAN based attack by Hitai et al., adversarial examples of images and text using GAN model by Zhao et al., variational autoencoder (VAE) – GAN, adversarial autoencoder (AAE), etc. (Duddu, 2018; Li et al, 2018).

*2.6.4 Self-Organized Maps (SOM)*

A neural-network based and unsupervised learning algorithm that was developed by Kohonen in 1982 (Tsai et al., 2009; Chakraborty et al., 2018). The algorithm uses a process of self-organization and reduces the dimensionality of input vectors into usually two dimensions representative output vectors (Tsai et al., 2009). The algorithm also uses the winning node which is the neuron closest to the training set and group the output vectors with similar weights in a self-organized ordered map after training (Tsai et al., 2009).

*2.6.5 Genetic Algorithm (GA) or Genetic Programming (GP)*

The algorithm belongs to the group of Evolutionary Computation (EC) which uses the biologically inspired natural selection and evolution process (Tsai et al., 2009; Nguyen, 2018). The algorithm generates a large population of variables of interest or candidate programs and evaluates the performance of those using the fitness of measure function (Elwahsh et al., 2018; Tsai et al., 2009). Any weak performing variables are replaced by high performing variables using the genetic recombinant, natural selection, crossover, and mutation in large iterations (Tsai et al., 2009; Nguyen, 2018). Other algorithm or technique such as Particle Swarm Optimization and Ant Colony Optimization belong to EC group (Nguyen, 2018).

### *2.6.6 Fuzzy logic*

Fuzzy logic is based on the Fuzzy set theory and explains the phenomenon in real-world using the values for reasoning that are between 0 and 1 (Tsai et al., 2009). The degree of logic is flexible and not a hard statement of true or false (Tsai et al., 2009). For example, rain is a natural phenomenon and its chances and amount falling varies (Tsai et al., 2009).

## 3. Adversarial security attacks on machine learning

### *3.1 Brief overview*

A variety of taxonomies, definitions, models, and categories of security attacks on machine learning (ML) and deep learning (DL) systems that are based on the adversarial capabilities, knowledge, goals, resources, and the target criteria exists (Corona, Giacinto, & Roli, 2013; Li et al., 2018; Sethi & Kantardzic, 2018a) and some of which are presented below. According to Barreno (2008), Barreno et al. (2006 and 2010), Biggio and Roli (2018), Chakraborty et al. (2018), Li et al. (2018), Liu et al. (2018), Ozdag (2018), Shi, Sagduyu and Grushin (2017), and Shi and Sagduyu (2017), the adversarial security attacks of ML and DL systems can be described along the following schemes:

- Influence
  - Causative attack. Targets the training process or the training data is altered. The model trained on the altered data provides the manipulated output. It is sometimes also called the poisoning attack.
  - Exploratory attack. Targets after the training process. Explores or probe the learner for useful information. Can exploit misclassifications but do not alter the training process.
  - Evasion attack. Targets after the training process. Modifies the input data to the learner that results in an incorrect prediction or evade detection.
- Specificity
  - Targeted attacks. Targets the specific points, instances, or exploits that are continuous streams.
  - Indiscriminate. Targets the general class of points, instances, or exploits in a random non-targeted manner.
- Security violation
  - Integrity attack. A successful attack on assets via false negatives and that is being classified as normal traffic.
  - Availability attack. A broad class of an attack that makes the system unusable with classification errors, denial of service, false negatives and positives, etc.
  - Privacy violation attack. An exploratory attack type that reveals sensitive and confidential information from the data and models. Also known as model extraction, inversion, or hill-climbing attack.

Also, according to Biggio and Roli (2018), Chakraborty et al. (2018), Duddu (2018), Li et al. (2018), Yuan (2018), the adversaries might have different capabilities, goals, and a complete or partial knowledge of ML and DL systems with the phases of data input, features selection, algorithms and parameters, training the model and output. They explained the adversarial security environment using the following taxonomies and definitions:

- White box attack. An adversary has a complete knowledge of the ML and DL models or systems.
- Gray box attack. An adversary has some knowledge of the ML and DL models or systems.
- Black box attack. An adversary has no knowledge of the ML and DL systems. They are classified into non-adaptive, adaptive, and strict black box.
- Targeted and untargeted Attack. A targeted security attack targets a targeted class of an input or output. An example: intrusion detection binary classification system. The untargeted (or indiscriminate) does not target a targeted class of an input or output. An example: a single image perturbation.
- Adversarial capabilities. Data injection, modification, and logic corruption.
- Adversarial goals. Confidence reduction, misclassification, target misclassification, source and target misclassification.
- Attack frequency. One-time attacks; iterative attacks
- Transferability property. The lack of knowledge of the target model can be compensated by the local trained model of the attacker. The adversarial examples that mislead model A are likely to mislead model B. Also, the adversarial examples can be transferred or used on different models with different algorithms.
- Training phase modification: Adversaries can fine-tune parameters and manipulate labels and inputs.
- Testing phase modification: Using a white-box setup, adversarial sample crafting using the sensitivity estimation and perturbation selection.
- Privacy preserving models: Ensures the privacy of user’s data - CryptoNet, Deep Learning using garbled circuits (GC), etc. and use randomization algorithms, secure multi-party computation, homomorphic encryption (HE), etc.

Finally, there are some additional terms and definitions that aid in understanding the adversarial security attack scenarios.

- Adversarial Examples (AE). An image is vulnerable to noise introduction, natural perturbations (such as fast gradient sign, fog, and sunlight, etc.) or pixels manipulation (Ozdog, 2018) and that can be exploited by the adversary. The adversarial examples or attacks are also “subtly altered images, objects, or sounds that fool AIs without setting off human alarm bells,” explained Hutson (2018; para 4). Several benchmark datasets are available or real-world inputs that can serve as adversarial examples. A classifier can be fooled into misclassifying the (perturbed) sample which possibly can still be classified by a human eye (Ozdog, 2018). There are targeted and untargeted or non-targeted attacks using adversarial examples and those are based on perturbations according to some distance metrics (Wang, 2018).
- Perturbations. Disturbances or changes (or adding some noise) in the appearance of an image or motion of an object are called perturbations. Small perturbations that are close to the original samples and unnoticeable by a human eye, is the basis for adversarial examples that are used for training datasets or observations of the security attacks in the adversarial environment (Yuan et al., 2018). A small perturbation or minimally perturbed adversarial example (an example case of evasion attack); however, is not always good in the security assessment of ML or DL algorithm; rather, it is good for observing the sensitivity of the algorithm, reported Biggio and Roli (2018). For a thorough security assessment, the parameters related to maximum confidence of the classifier should be considered, because the adversary will try to maximize that on the output (Biggio & Roli, 2018). Those perturbations affect the performance of ML and DL models and they are classified as follows: perturbation scope (individual or universal); perturbation limitation (optimized or constrained); perturbation measurement ($l_p$, the magnitude of the perturbation by the p-norm distance; $l_0$, $l_2$, $l_\infty$ $l_p$ metrics) or psychometric perceptual adversarial similarity score (PASS)) (Yuan et al., 2018). Also, the methods of one-shot/one-step and iterative are used for generating adversarial perturbations that involve using a single and multiple steps computation (Akhtar & Mian, 2018). Finally, the amount of perturbations and its effectiveness on the successfulness of attack or the effect on the classifier's confidence level can also be described using the best-effort attack (a minimum amount of perturbation) and bounded attack (a fixed-range of perturbations) (Li et al., 2018).

- Distance Metrics. The measurements of the similarity between an adversarial example and its original copy (Wang, 2018). There are three widely used distance metrics in the $L_p$ form such as $L_1$, $L_2$, and $L_3$ (Li et al., 2018; Wang, 2018).
- Benchmark datasets. The availability of small or large datasets, well-known ML and DL models in the public domain, and the complex and computationally intensive learning models make it hard to attack and defend for both adversaries and defenders (Yuan et al., 2018). There are some well-known datasets such as ImageNet, MNIST, CIFAR, GTSRB, etc. and models such as AlexNet, GoogLeNet, LeNet, VGG, CaffeNet, ResNet, etc., useful in evaluating adversarial security attacks (Yuan et al., 2018).
- Defenses. There are different defensive mechanisms and approaches to adversarial security attacks (Carlini et al., 2019). For example, the security assessment mechanisms, countermeasures in the training phase, countermeasures in the testing phase, data security and privacy, according to Liu et al. (2018). And the modified input, modified networks, network add-ons, etc., according to Akhtar and Mian (2018). Brute-force adversarial training, data compression and randomization, gradient regularization, defense distillation, DeepCloak, Reject on Negative Impact (RONI), etc. are some examples of defenses (Liu et al., 2018; Akhtar & Mian, 2018).

### *3.2 . Adversarial security attacks and perturbations*

#### *3.2.1 Poisoning attacks*

Poisoning attacks aim to misclassify samples or output test data using the poisoned samples or adversarial examples in the training data set (Biggio & Roli, 2008; Duddu, 2018). The attack vector is exclusive to training data set and depending on the attacker capabilities, maximizes ML and DL models' classification or clustering errors, according to Duddu (2018) and Gardiner and Nagaraja (2016). Some examples of poisoning attacks include: boiling frog attack (An iterative attack where test data is poisoned incrementally and over time), label-flipping attack (A causative integrity/availability attack where an attacker introduces noise or flips labels (legitimate to malicious or vice-versa) in the training data set), bridging attack (a causative integrity/availability attack that introduces points between clusters, making them to join or split), etc. (Duddu, 2018; Gardiner & Nagaraja, 2016).

*3.2.2 Anchor points (AP) attacks*

Attackers with a limited probing budget, malicious samples, and goals of evasion and immediate benefits to them exploit the anchor points attacks (Sethi & Kantardzic, 2018a). In these attacks, the legitimate samples serve as anchor points or seed samples that further allows to advance attacks or explore the model workings (Sethi & Kantardzic, 2018a). The zero-day exploits of the vulnerabilities without their fixes is one example (Sethi & Kantardzic, 2018a).

*3.2.3 Reverse engineering attacks (RE) attacks*

Reverse engineering attacks avoid detection, extract features related to the classifiers, and obstruct retraining processes (Sethi & Kantardzic, 2018a). The selection of models, attack environment, scalability of the data, and availability of probes, etc. influence the reverse engineering attacks (Sethi & Kantardzic, 2018a). These types of attack require both the legitimate and malicious sample to be successful, whereas in comparison to anchor points (AP) that require only the legitimate seed sample (Sethi & Kantardzic, 2018a).

*3.2.4 Dictionary attacks*

Dictionary attacks are a form of poisoning attacks and are categorized into indiscriminate or targeted type (Gardiner and Nagaraja, 2016). They belong to the family of causative attacks and work against the classifiers train on words or token-based features (Gardiner and Nagaraja, 2016). Attackers insert malicious data with an intention that it will be included in future training data sets and cause classification (Gardiner and Nagaraja, 2016). For example, SpamBayes - a classifier system - that is fooled by spam email containing legitimate dictionary words, as cited in Gardiner and Nagaraja (2016).

*3.2.5 Mimicry attacks*

Mimicry attacks are the exploratory attacks type and where the attackers use the attack points that appear to be a benign point (Gardiner and Nagaraja, 2016). The attacks focus on exploring benign and malicious points in the data features, and not acquiring explicit knowledge about the classification algorithm (Gardiner and Nagaraja, 2016). The attacks become successful by reducing the distance between the attack points and benign points in a targeted and indiscriminate manner (Gardiner and Nagaraja, 2016). These attacks mostly have demonstrated against the classifiers but clustering algorithms with distance functions are also vulnerable (Gardiner and Nagaraja, 2016). Wagner and Soto (2002) introduced these attacks, as cited in Gardiner and Nagaraja (2016).

*3.2.6 Equation-solving or model extraction attacks*

Equation-solving attacks target models with equations and their variables (Duddu, 2018; Nguyen, 2018). The attackers can plug-in values of the variables and find the values of the unknown variables; thereby, revealing some information about the models and the architectures to the attackers (Duddu, 2018). The popular models such as neural networks, decision trees, multilayer perceptron (MLP), variants of logistic regression - binary logistic regression (BLR) and multiclass logistic regression (MLR) - and APIs (application programming interface) within the cloud-based machine learning (ML) services (such as BigML, Amazon ML, etc.) are the most targeted and vulnerable to equation-solving attacks (Duddu, 2018; Chakraborty et al., 2018; Nguyen, 2018). Tramer et al. demonstrated some examples of this black-boxed, model extraction attack (Chakraborty et al., 2018; Nguyen, 2018).

*3.2.7 Model inversion attacks*

Model inversion attacks allow the attackers to gain insight about the training data and further aid in cloning the model that is close to the actual model (Nguyen, 2018). These types of attacks exploit the confidence values of the output of the model in a variety of settings and applications (Chakraborty, 2018). Some demonstrated examples of this attack type can be seen with models based on multilayer perceptron (MLP), denoising auto-encoder (DAE) network, SoftMax regression and Machine learning (ML) APIs (Chakraborty et al., 2018). Fredrickson et al. demonstrated these attacks using their proposed algorithm using the least biased maximum a posteriori (MAP) for the computation of the optimal input features (Duddu, 2018; Nguyen, 2018).

*3.2.8 Model or membership inference attacks*

Model inference attacks involve the attacker sending well-crafted queries to the target models and obtaining their predictions (Chakraborty et al. 2018; Duddu, 2018). The attackers may employ shadow models for generating queries that target attack models and in turn gain insight on the memberships or classes of those queries belonging to the target dataset (Chakraborty et al. 2018). There are demonstrated examples of these attacks involving Markov models, public APIs, and Google and Amazon's online Machine learning (ML) services (Chakraborty et al. 2018).

*3.2.9 Path-finding attacks*

Pathfinding attacks explore or traverse the trees such as binary trees, multi-nary trees, and regression trees for the path until the end node or leaf is reached (Duddu, 2018). The attackers try to manipulate the input features and repeat the process until successfully finding the next node or leaf (Nguyen, 2018). Tremor et al. proposed pathfinding attacks and use parameters of a tree, node, and an identifier (Nguyen, 2018).

3.2.10 Least cost and constrained attacks

Lease cost and constrained attacks require a trade-off between the minimum perturbation and the successfulness (or the effectiveness) of the attack (Li et al., 2018). The attack success is measured by the impact on the classifier's confidence by the perturbed differences between the distorted and original input (Li et al., 2018). In the least cost and constrained context, the attacker would require the minimum amount of perturbations and is also restricted by the constrains of target class or box-constraint. Szegedy et al. successfully demonstrated these types of attacks (as cited in Li et al., 2018).

### 3.2.11 Causative integrity attacks

- Red herring attacks. Red Herring attacks, introduced by Newsome et al. in 2005 and 2006, exploit weaknesses in the conjunction learners of the polymorphic virus detector (Barreno et al., 2010). The polymorphic virus detector thwart security attacks using virus signatures and those are learned using a conjunction learner and Naïve-Bayes learner (Barreno et al., 2010). The attacker uses training distribution data set used by the defender and introduce malicious features, bypassing the detector (Barreno et al., 2010).
- Prove Approximately Correct (PAC) learning framework attacks. These types of attacks are concerned with bounds of malicious errors in the training data (Barreno et al., 2010). Kearns and Li extended the work of Valiant's PAC learning framework and proved that the attacker can restrict the confidence of the learners or the success of an algorithm, given by the probability of incorrect prediction (Barreno et al., 2010). These attacks can be targeted and indiscriminate and fits the description of a causative game (Barreno et al., 2010).

### 3.2.12 Causative availability attacks

- Correlated outlier attacks. Correlated outlier attacks, introduced by Newsome et al in 2006, exploit weakness in the naïve bays like learner of the polymorphic virus detector. The attackers add malicious features in the positive instances of the training data set and cause the detector to block the normal traffic, much like a denial of service attack. These attacks can be targeted and indiscriminate and fits the description of the cognitive game (Barreno et al., 2010).
- Allergy attacks. Allergy attacks, introduced by Chung and Mok in 2006 and 2007, attack the autograph warm signature generation system (Barreno et al., 2010). The signature system works by analyzing the infected nodes and malicious traffic using the behavioral patterns and blocking rules (Barreno et al., 2010). The attacker can manipulate the system by scanning the network and flood the crafted packets toward the targeted nodes, which results in legitimate traffic being blocked as a DoS prevention (Barreno et al., 2010). These attacks fit the description of a causative game (Barreno et al., 2010).

### 3.2.13 Exploratory integrity attacks:

- Polymorphic blending attacks. In the polymorphic blending attacks, the attackers encrypt their traffic which closely resembles the normal traffic in order to evade intrusion detection systems (IDS). Foogla and Lee in 2006 demonstrated the example of these attacks in 2006 (Barreno et al., 2010).
- Sequence based intrusion detection system (IDS) attacks. A type of mimicry attack that uses the modified version of exploits such as the password and trace out programs, sequences of system calls against the IDS, as demonstrated by Tan et al. (2002), and a framework such as pH for validating vulnerabilities of IDS, as demonstrated by Wagner and Soto (2002) (Barreno et al., 2010).

- Good word attacks. In the good word attacks, the attackers add good words or non-spam words in the spam emails and evade the detection by the spam filters (Barreno et al., 2010). Lowd and Meek (2005b) and Wittel and Wu (2004) demonstrated the examples of these (as cited in Barreno et al., 2010).
- Reverse engineering classifiers attacks. These types of attacks consider a cost function rather than the positive and negative labels of the classifier (Barreno et al., 2010). An algorithm is already provided to the attackers for reverse engineering classifiers, as demonstrated by Lowd and Meek (2005a) (as cited in Barreno et al., 2010).

### *3.2.14 Exploratory availability attacks*

Any types of denial of service attacks fall into this category and are common for non-learning systems. However, they are not so common for the learning processes or systems. Some demonstrated examples are there such as the attacker taking advantage of the computationally expensive image processing to scan advertisements in the spam detection systems (Dredze, Gevaryahu, & Elias-Bachrach, 2007; Wang et al., 2017), or the attacker convincing the IPS trained on intrusion traffic to block the legitimate host or drop the normal traffic (Barreno et al., 2018).

### *3.2.15 Stackelberg prediction game (SPG) attacks or game-theoretical perspective*

In these types of attacks a game scenario is depicted where the attackers attack the learning models and try to prevent learning new knowledge or feed garbage knowledge that results in misclassification. The attack strategies are to maximize the cost function of the model. In the case of different attacking goals and attackers, Zhou and Kantarcioglu described the attacks as Nested Stackelberg Game attacks (as cited in Nguyen, 2018). Those can be prevented using the set of models to confuse the attackers as one of the strategies (Nguyen, 2018). A cost-sensitive game-theoretical perspective has also been used for adversarial classification by Dalvi et al. (as cited in Chakraborty et al., 2018).

### *3.2.16 Gradient descent (GD) attacks*

Gradient descent attacks are common in both the supervised and unsupervised learning and rely on the detailed model information (Gardiner & Nagaraja, 2016). Some examples of that include Fast-Gradient Sign Method (FGSM), DeepFool, Jacobian Saliency Map Attack (JSMA), Houdini, Carlini-Wagner (CW) attack, Basic Iterative Method (BIM), etc. (Brendel, Rauber, & Bethge, 2018). GD is an optimization algorithm that seeks to minimize the output function and moving towards the steepest descent (Gardiner & Nagaraja, 2016). The attackers generate attack points which are then tested for their effectiveness using their own learners, and in the case of non-effectiveness, continue to generate new points using GD until the attacks are successful (Gardiner and Nagaraja, 2016). One defense approach is to mask the gradients and techniques and means such as defensive distillation, saturated nonlinearities, and nondifferentiable classifiers exist for that (Brendel, Rauber, & Bethge, 2018; Papernot et al., 2016c).

3.2.17 Fast gradient sign method (FGSM)

Fast gradient sign method is used to generate adversarial examples and was introduced by Goodfellow et al. (Duddu, 2018). It is among the very first and few methods to test using the adversarial examples against the against the neutral networks and deep learning models (Li et al., 2018). FGSM is quite fast in operations because it takes only the gradient information (Li et al., 2018). FGSM calculates the gradient of the cost function using the input features or perturbations (Chakraborty et al., 2018). FGSM has been tested on GoogLeNet images with a small input feature vector. The elements of the vector are comprised of the sign of the gradient of the loss function (or the cost function), the perturbation, parameters, and input (Ozdag, 2018). There are various variants of FGSM exist such as the target class method, basic iterative method (BIM), projected gradient descent (PGD), L-BFGS, etc. (Chakraborty et al., 2018; Li et al., 2018; Ozdag, 2018).

*3.2.18 Carlini-Wagner (CW) attacks*

An adversarial example technique that is capable of evading various defenses including the recent defensive distillation (Duddu, 2018; Ozdag, 2018). CW uses three types of attack: $L_2$ attack, $L_0$ attack, $L_\infty$ attack and beat some of the promising defense algorithms. Defensive distillation which was developed to test the robustness of the neutral network is also vulnerable to CW attacks (Papernot et al., 2016c; Carlini et al., 2019). $L_0$ was the first technique used in misclassification using the perturbed images of ImageNet. $L_0$ attack measures the distance between the given input and the perturbed input, whereas $L_2$ attack measures the root mean score between the given input and the perturbed input. Similarly, $L_\infty$ measure the maximum changes in all dimensions. One example of CW attack is demonstrated by Chen et al. (2017) (as cited in Rauber & Bethoe, 2018).

*3.2.19 Jacobian-based saliency MAP (JSMA) attacks*

This approach was proposed by Papernot et al. and uses adversarial saliency scores to identify the sensitivity of the model or fool the network (Duddu, 2018; Ozdag, 2018). The approach iteratively uses input features for perturbations and construct a map of those relating to output variations. This method is called a forward derivative and uses a matrix defined as the Jacobian of the function or Jacobian of the training model (Chakraborty et al., 2018; Ozdag, 2018). The saliency map is like a visualization tool that allows to generate and explore adversarial examples that causes the desired changes in the classifier's output (Li et al., 2018; Ozdag, 2018). One example of such an attack include the black box variants of JSMA by This includes black-box variants of JSMA by Narodytska and Kasiviswanathan (2016) (as cited in Brendel, Rauber, & Bethge, 2018)

## 4. Comparisons of related work

A comprehensive list of security attacks and perturbations and their related reference studies is provided in the table 1 in Appendix A. Below are the brief comparisons of three most common security attacks and perturbations and the preferred ML/DL methods and datasets. A study of intrusion detection by machine learning by Tsai et al. (2009) inspired this comparison work and much of the presentation style has been adopted from the same reference.

## 4.1 Security attacks and perturbations

The three most common security attacks and perturbations are chosen based on the largest number of counts in that category from the table 4 in Appendix A. The resulting Table 1 generated, reveals the following: Poisoning / Causative (18); Adversarial Example / Evasion (62); and Exploratory (13). Figure 1 presents a column chart of the yearwise distribution of reference studies for security attacks / perturbations. The chart reveals a majority of the work have been performed in the last three years and the trend appears to be positive and upward for the all three categories. We predict a similar trend for all the categories in future.

## 4.2. Machine learning and deep learning methods

A variety of machine learning and deep learning methods, pre-trained models and systems have been used to study adversarial security attacks and perturbations. The methods of SVM, NN, and DNN have been the popular choices. Their growth has been steady in the last four years and most of the work on those has been performed in the last two years. Other methods such as CNN, DT, NB, LR, etc. have also been the popular choices for the studies.

## 4.3 Datasets

Table 3 presents the three most popular datasets and their reference studies. Figure 3 presents a column chart of the yearwise distribution of the datasets with their matching reference studies. MNIST, CIFAR, ImageNet have been the popular choices of datasets and their trend is growing yearwise for security attacks and perturbations. Some other popular choices of datasets also include GTSRB (German Traffic Sign Recognition Benchmark), Contagio, Synthetic Images, Real-World datasets, etc.

**Table 1**

Total number of reference studies for the most common security attacks and perturbations

| Security Attacks / Perturbations | Poisoning / Causative | Adversarial Example / Evasion | Exploratory |
|---|---|---|---|
| No. of Reference Studies | 18 | 62 | 13 |
| Reference Studies | Baracaldo et al. (2018); Barreno (2008); Biggio (2016); Biggio et al. (2014); Biggio, Nelson, and Laskov (2013); Chen et al. (2018); Jagielski et al. (2018); Kloft and Laskov (2012); Kumar and Mehta (2018); Li et al. (2016); Mozaffari-Kermani et al. (2015); Nelson et al. (2008); Rubinstein et al. (2009); Shi and Sagduyu (2017); Suciu et al. (2018); Xiao (2012); Xiao, Xiao, and Eckert (2012); Yang et al. (2017) | Alfaro (2018); Athalye et al. (2018); Bhagoji et al. (2017a, b); Biggio (2016); Biggio et al. (2017); Chen et al., 2017; Cisse et al. (2017); Comesaña, Pérez-Freire, & Pérez-González (2006); Czaja et al. (2018); Elsayed, Goodfellow, and Sohl-Dickstein (2018); Eykholt et al. (2018a); Eykholt et al. (2018b); Goodfellow et al. (2014); Goodfellow, Shlens, and Szegedy (2015); Han and Rubinstein (2017); Hayes & Danezis (2017); Jia and Gong (2018); Kantchelian, Tygar, & Joseph (2016); Khalid et al. (2018a, b); Khalid et al. (2019); Kloft and Laskov (2012); Kolosnjaji et al. (2018); Kos, J., Fischer, and Song (2017); Kurakin, Goodfellow, & Bengio (2017); Kurakin, Goodfellow, and Bengio (2017); Larsen et al. (2016); Levina, Sleptsova, Zaitsev (2016); Li & Vorobeychik (2014); Liang et al. (2017); Liu et al. (2017); Liu et al. (2018a); Lowd, D., & Meek., C. (2005); Luo et al. (2018); Madry et al. (2017); Melis et al. (2018); Moosavi-Dezfooli, Fawzi, & Frossard (2016); Narodytska and Kasiviswanathan (2016); Norton and Qi (2017); Ozdag (2018); Papernot et al. (2016); Papernot et al. (2017); Papernot, McDaniel, and Goodfellow (2016); Quiring, Arp, and Rieck (2018); Shi and Sagduyu (2017); Sitawarin et al. (2018); Srndic and Laskov (2014); Su, Vargas, and Kouichi (2017); Suciu et al. (2018); Suya et al. (2017); Szegedy et al. (2014); Wang et al. (2017); Xiao (2012); Xiao et al. (2015); Xiao et al. (2017); Xiao, Xiao, and Eckert (2012); Xu, Evans, and Qi (2018); Xu, Qi, and Evans (2016); Yakura and Sakuma (2018); Zhang et al. (2016) | Ateniese et a. (2015); Barreno (2008); Nguyen (2018); Salem et al. (2018); Sethi and Kantardzic (2018a); Sethi, Kantardzic, and Ryu (2018); Shi and Sagduyu (2017); Shokri et al. (2017); Tramer et al. (2016); Wang and Gong (2018); Xue and Chuah (2018); Yeom et al. (2018) |

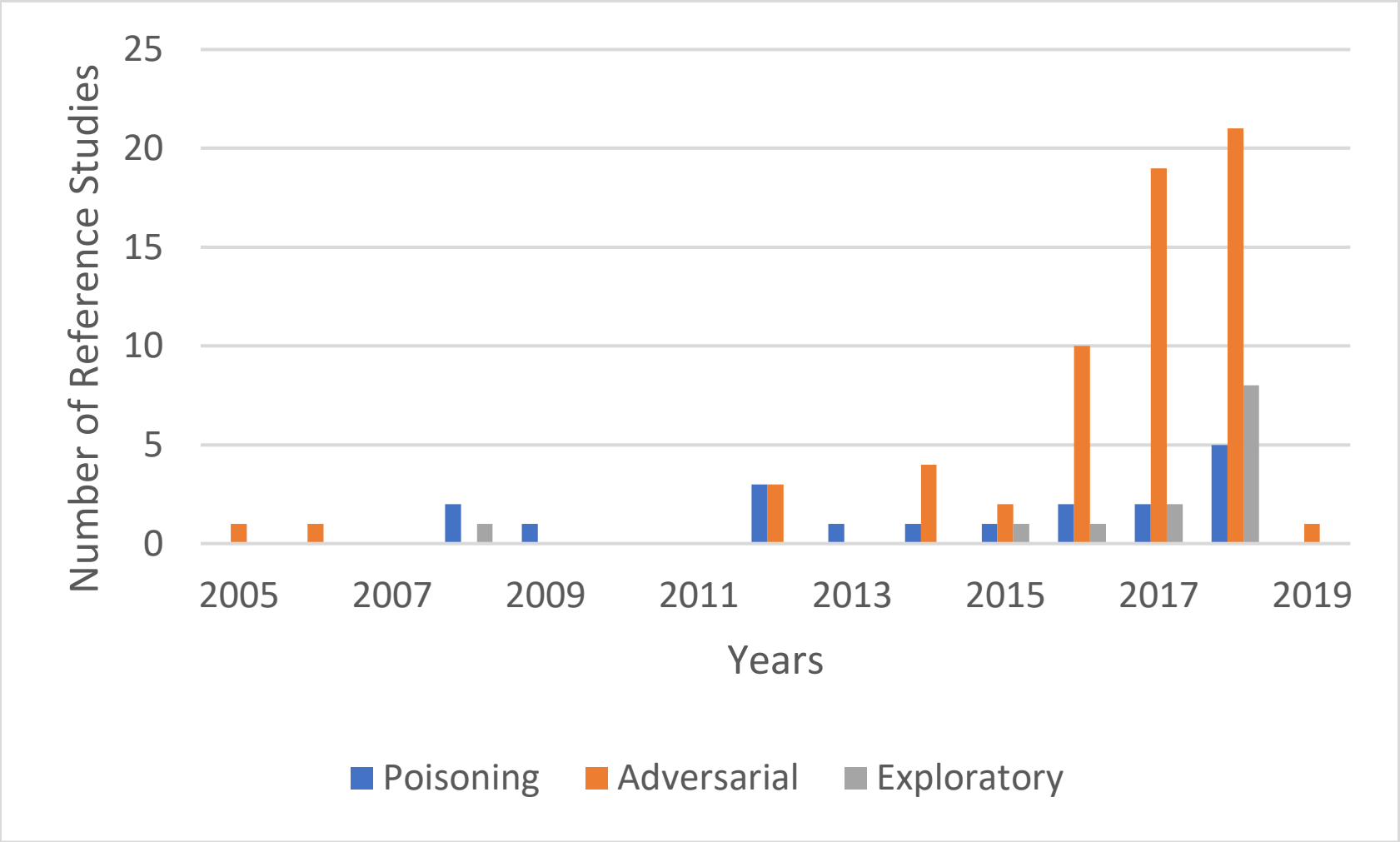

**Figure 1.** Yearwise distribution of reference studies for security attacks / perturbations

**Table 2**

Total number of reference studies for the most common ML/DL methods

| ML/DL Methods | SVM | NN | DNN |
|---|---|---|---|
| No. of Reference Studies | 34 | 23 | 20 |
| Reference Studies | Chen et al. (2018); Biggio, Nelson, and Laskov (2013); Baracaldo et al. (2018); Biggio et al. (2014); Xiao et al. (2015); Xiao (2012); Xiao, Xiao, and Eckert (2012); Liu et al. (2018a); Shi, Sagduyu, and Grushin (2017); Sethi and Kantardzic (2018a); Sethi, Kantardzic, and Ryu (2018); Levina, Sleptsova, Zaitsev (2016); Biggio et al. (2017); Srndic and Laskov (2014); Ateniese et a. (2015); Papernot, McDaniel, and Goodfellow (2016); Zhang et al. (2016); Wang and Gong (2018); Melis et al. (2018); Bhagoji et al. (2017a, b); Suya et al. (2017); Han and Rubinstein (2017); Bulò et al. (2017); Kantarcioglu and Xi (2017); Kantarcioglu and Xi (2016; 2017); Zhou and Kantarcioglu (2016); Nguyen (2018); Zhang and Zhu (2018); Bruckner and Scheffer (2009); Bruckner and Scheffer (2011); Nguyen (2018); Vorobeychik and Li (2014); Li & Vorobeychik (2014) | Alfaro (2018); Athalye et al. (2018); Athalye, Carlini, and Wagner (2018); Biggio et al. (2017); Carlini and Wagner (2017); Chen et al. (2017); Clements and Lao (2018); Fredrikson, Jha, Ristenpart (2015); Han and Rubinstein (2017); Hanzlik et al. (2018); Hayes & Danezis (2017); Jia and Gong (2018); Kolosnjaji et al. (2018); Kumar & Mehta (2018); Kurakin, Goodfellow, and Bengio (2017); Madry et al. (2017); Oh et al. (2018); Ozdag (2018); Shokri et al. (2017); Suciu et al. (2018); Szegedy et al. (2014); Wang and Gong (2018); Yang et al. (2017) | Danezis (2017); Ji et al. (2018); Khalid et al. (2018a, b); Khalid et al. (2019); Liang et al. (2017); Liu et al. (2017); Luo et al. (2018); Nguyen (2018); Ozdag (2018); Papernot et al. (2016); Papernot et al. (2017); Papernot, McDaniel, and Goodfellow (2016); Quiring, Arp, and Rieck (2018); Su, Vargas, and Kouichi (2017); Tramer et al. (2016); Wang and Gong (2018); Wang et al. (2017); Xu, Evans, and Qi (2018); Yakura and Sakuma (2018); Yang et al. (2017) |

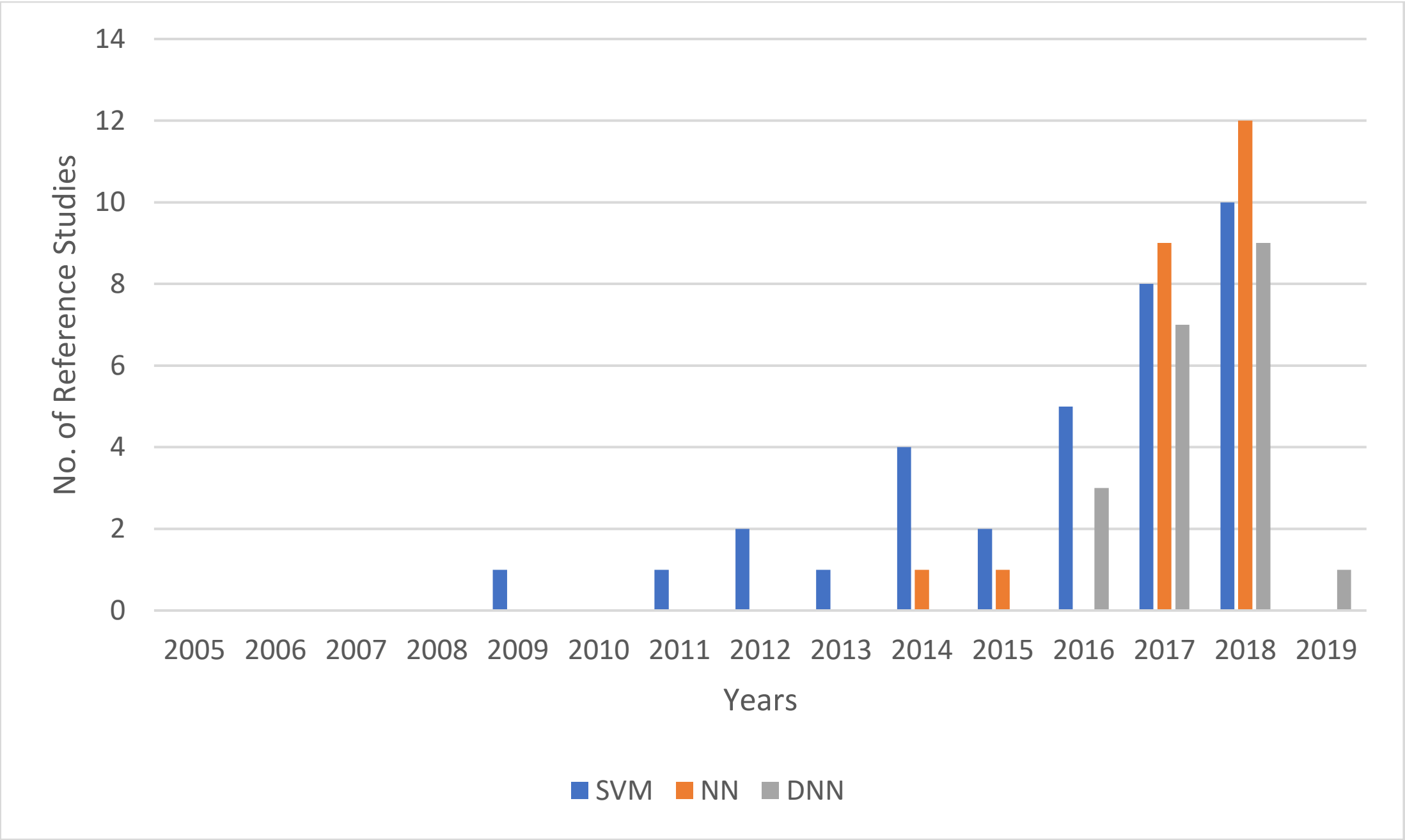

**Figure 2.** Yearwise distribution of reference studies for ML/DL methods

**Table 3**

Total number of reference studies for the most common datasets

| Datasets | MNIST | CIFAR | ImageNet |
| --- | --- | --- | --- |
| No. of Reference Studies | 36 | 22 | 11 |
| Reference Studies | Alfaro (2018); Athalye, Carlini, and Wagner (2018); Biggio et al. (2014); Biggio et al. (2017); Biggio, Nelson, and Laskov (2013); Brendel, Rauber, and Bethge (2018); Carlini and Wagner (2017); Chen et al. (2017); Chen et al. (2017b); Clements and Lao (2018); Elsayed, Goodfellow, and Sohl-Dickstein (2018); Engstrom, Ilyas, and Athalye (2018); Goodfellow et al. (2014); Goodfellow, Shlens, and Szegedy (2015); Han and Rubinstein (2017); Hanzlik et al. (2018); Kantchelian, Tygar, & Joseph (2016); Kos, J., Fischer, and Song (2017); Luo et al. (2018); Madry et al. (2017); Moosavi-Dezfooli, Fawzi, & Frossard (2016); Narodytska and Kasiviswanathan (2016); Norton and Qi (2017); Oh et al. (2018); Ozdag (2018); Papernot et al. (2016); Papernot et al. (2017); Papernot, McDaniel, and Goodfellow (2016); Salem et al. (2018); Shokri et al. (2017); Szegedy et al. (2014); Wang et al. (2017); Xiao et al. (2017); Xu, Evans, and Qi (2018); Yang et al. (2017); Zhang and Zhu (2018) | Athalye, Carlini, and Wagner (2018); Brendel, Rauber, and Bethge (2018); Carlini and Wagner (2017); Chen et al. (2017); Chen et al. (2017b); Clements and Lao (2018); Elsayed, Goodfellow, and Sohl-Dickstein (2018); Engstrom, Ilyas, and Athalye (2018); Goodfellow et al. (2014); Hanzlik et al. (2018); Hayes & Danezis (2017); Luo et al. (2018); Madry et al. (2017); Moosavi-Dezfooli, Fawzi, & Frossard (2016); Narodytska and Kasiviswanathan (2016); Ozdag (2018); Salem et al. (2018); Shokri et al. (2017); Su, Vargas, and Kouichi (2017); Suciu et al. (2018); Xu, Evans, and Qi (2018); Yang et al. (2017) | Athalye, Carlini, and Wagner (2018); Brendel, Rauber, and Bethge (2018); Engstrom, Ilyas, and Athalye (2018); Hayes & Danezis (2017); Ji et al. (2018); Kurakin, Goodfellow, & Bengio (2017); Oh et al. (2018); Ozdag (2018); Su, Vargas, and Kouichi (2017); Szegedy et al. (2014); Xu, Evans, and Qi (2018) |

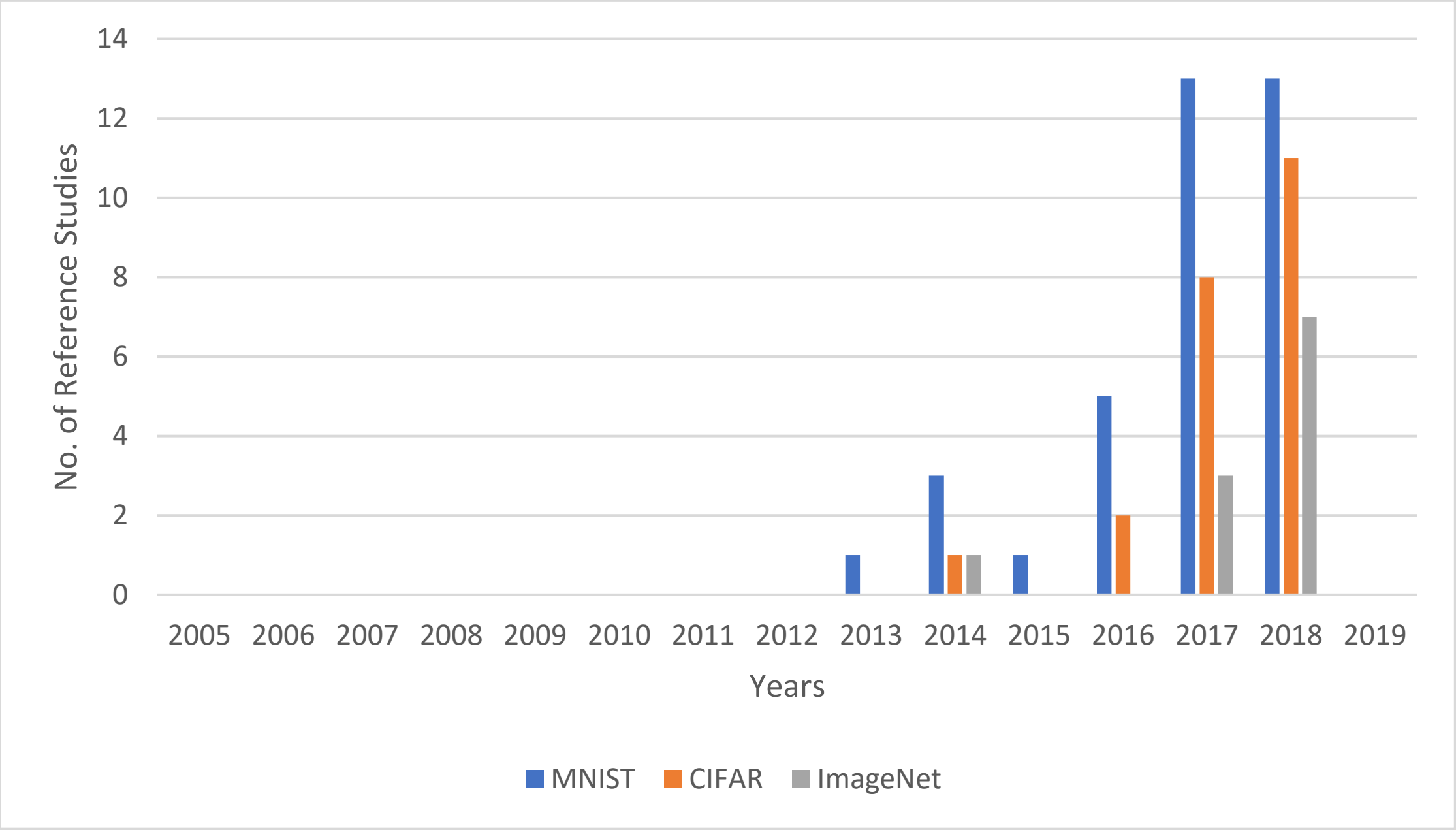

**Figure 3.** Yearwise distribution of reference studies for the datasets

## 5. Discussion and conclusion

The machine learning (ML) and deep learning (DL) methods offer promising solutions to services and products. The underlying technologies of these methods, however, are new and still evolving. The development and implementation of DM and ML models themselves have been a challenge as they require expert knowledge from multiple domains. Depending on the requirements and goals, the models are continuously improved for their accuracies manipulating the parameters of equations and for their performances deploying in the distributed systems or cloud environments. The source of training and test data and the deployed environment are also sometimes guarded for privacy and secrecy of the learning models. All these different scenarios and factors contribute to the vulnerabilities of the ML and DL methods and those can be exploited by the adversaries. In this paper, we presented some simplified knowledge pertaining to ML and DL methods and relevant adversarial security attacks and perturbations. Specifically, we focused on the background, definitions, terms, descriptions, and comparisons of ML and DL methods and security attacks. A comprehensive list of security attacks and perturbations is provided in the table in Appendix A. We hope the new entrants of cyber security in the research area will benefit tremendously from the new and structured knowledge offered in this paper. Even though several research papers exist that reviews adversarial security attacks and perturbations, but there is always room to grow due to the dynamic nature of ML and DL methods. The learning models that are ideal and produce satisfying results remain an open and a lasting challenge. This includes the issue of adversarial security attacks and perturbations because of its relation to the DM and ML methods. Also, a few papers cover the defenses of adversarial security attacks and perturbations; however, some of those defenses have already been beaten or their robustness are questionable (Carlini et al., 2019). In our future work, we anticipate covering those in a revised version or another series of this paper.

# Appendixes

## Appendix A

Table 1
A list of adversarial security attacks and perturbations

| Adversarial Security Attacks & Perturbations | Types, Features or Target | Methods/Approaches | ML and DL Methods Involved | Data Used | Reference Studies |
|---|---|---|---|---|---|
| Poisoning | Training Data | Mobile Malware Detection Systems | SVM, RF, KNN | Android Application Samples | Chen et al. (2018) |
| Poisoning | Gradient-Ascent | Gradient Ascent Strategy | SVM | MNIST | Biggio, Nelson, and Laskov (2013) |
| Poisoning | Evasion | General Framework; Biometric Identity Recognition System | Classifiers | N/A | Biggio (2016) |
| Poisoning | Training Data | Collaborative Filtering Algorithms | Projected Gradient Ascent (PGA) | MovieLens dataset | Li et al. (2016); Kumar and Mehta (2018) |
| Poisoning | Training Data | A Proposed Attack Procedure | Many | Healthcare Datasets. | Mozaffari-Kermani et al. (2015) |
| Poisoning | Training Data | Provenance Frameworks | SVMs | IoT Data | Baracaldo et al. (2018) |
| Poisoning | Evasion; Privacy Preserving | Prediction Errors | SVM | MNIST; Contagio Dataset | Biggio et al. (2014) |
| Label Noise | Adversarial Examples; Training Data | Security of SVM algorithms | SVM | Real-World Datasets at LibSVM website; Synthetic | Xiao et al. (2015) |
| Pattern Classifiers | Training and Testing Data Sets | A Framework for Empirical of Evaluation of Classifier Security | Different Classifiers and Algorithms | Different Ones | Biggio, Fumera, and Roli (2014a) |
| Control-Flow Hijacking; Denial-of-service (DoS) | Evasion | Vulnerabilities in DL Frameworks such as Caffe, Tensorflow, Torch | DL Frameworks Dependencies: NumPy; OpenCV | MNIST | Xiao et al. (2017) |
| False Data Injection (FDI) | Power Systems | Multiple Linear Regression Models | Linear Regression | IEEE 24-Bus and 118-Bus Systems | Zhang et al. (2018) |
| Boundary | Decision-Based; Blackbox | Clarifai.com Brand Recognition Model. | Cloud-based Computer Vision API by Clarifai | MNIST; CIFAR; ImageNet | Brendel, Rauber, and Bethge (2018) |
| Label Flipping | Poisoning; Adversarial Example | Adversarial Label Flips on SVMs | SVM - Linear and RBF Kernel | LIBSVM Real-World Datasets | Xiao (2012); Xiao, Xiao, and Eckert (2012) |
| Targeted Adversarial Defenses | Adversarial Defenses | Adversarial Logit Pairing (ALP); Kannan, Kurakin, and Goodfellow Threat Model; Projected Gradient Descent (PGD) | ALP Trained Model | CIFAR-10; MNIST; ImageNet | Engstrom, Ilyas, and Athalye (2018) |
| Iterative Optimization-Based; GD-Based | White-Box | Obfuscated Gradients; ICLR 2018 Defenses | NNs | CIFAR-10; MNIST; ImageNet | Athalye, Carlini, and Wagner (2018) |
| Targeted Adversarial Defenses | White-Box; Adversarial Defenses | Pixel Deflection; High-level Representation Guided Denoiser; Projected Gradient Descent (PGD) | Defenses | BPDA; PGD | Athalye and Carlini (2018) |
| Adversarial Examples | Fast-Gradient Sign Method (FGSM); Physical-World Inputs | Inception v3 Image Classification; Basic Iterative Method (BIM) | NNs | A Cell-phone Camera as an Input | Kurakin, Goodfellow, and Bengio (2017) |
| Physical-World Inputs; Disappearance Attacks; Creation Attacks | Adversarial Examples; Impersonate; Spoofing | YOLO Lab Environment; Outdoor Experiments; Faster R-CNN; LISA-CNN; RP2 Algorithm | DNNs | Posters; Stickers | Eykholt et al. (2018a); Eykholt et al. (2018b) |
| Physical-World Inputs | White-Box; Adversarial Examples | Sign Embedding Attack; CV-Based Systems of AVS; Traffic Sign Recognition Model | CNNs | German Traffic Sign Recognition Benchmark (GTSRB) | Sitawarin et al. (2018) |
| Adversarial Texture Malware Perturbation Attacks (ATMPA) | A Framework | Malware Visualization Method | CNN, SVM, RF | Kaggle Microsoft Malware Classification Challenge (BIG 2015). | Liu et al. (2018a) |
| One Pixel | Adversarial Examples | Differential Evolution (DE) | DNNs | CIFAR-10; ImageNet | Su, Vargas, and Kouichi (2017) |
| Poisoning; Fast Statistical | Poisoning | Theoretically-Grounded Optimization framework | Linear Regression | Healthcare; Loan; House Price Datasets | Jagielski et al. (2018) |
| Reverse Engineering (RE); Model Extraction; Membership Inference | ML API | MLCapsule - An Offline Deployment of MLaaS; Intel's Software Guard Extensions (SGX), | NNs | MNIST; CIFAR; GTSDB | Hanzlik et al. (2018) |
| Dictionary; Focused | Causative | SpamBayes - Spam Email System; Indiscriminate and Targeted Attacks | K-fold cross-validation | Text Retrieval Conference (TREC) 2005 | Nelson et al. (2008) |
| Multi-Agent Networks | Learning Rules | Byzantine Fault-Tolerant Non-Bayesian Learning (BFL) | Non-Bayesian Learning | N/A | Su and Vaidya (2016) |
| Adversarial Example | Evasion; White-box | Text-based; The Cost Gradients of the Input | DNN | Hot Training / Sample Phrases | Liang et al. (2017) |
| Physical | Adversarial Examples | Remote Sensing; Satellite Image Classification Problems | CNN-I (a variant of DenseNet) | Functional Map of the World (fMoW) | Czaja et al. (2018) |
| Exploratory; Causative; Evasion | Influence Attacks; Training Data | Text/Image Classification; CNTK | FNN | Attacker Test Data | Shi and Sagduyu (2017) |
| Black-Box | Exploratory; Training/Testing Data | CNTK; NTLK; Sci-Kit; Text Classification | NB; SVM; DL | Reuters-21578 | Shi, Sagduyu, and Grushin (2017) |
| Anchor Points (AP) | Exploratory; Blackbox | Seed-Explore- Exploit framework | KNN, SVM-RBF, DT, RF | KDD99, CAPTCHA, Spambase, etc. | Sethi and Kantardzic (2018a) |
| Anchor Points (AP); Reverse Engineering | Exploratory; Blackbox | Exploration-exploitation Based Strategy; A Data Driven Framework | KNN, SVM-RBF, DT, RF | KDD99, CAPTCHA, Spambase, etc. | Sethi, Kantardzic, and Ryu (2018) |
| Side-channel; Cryptoalgorithms | Evasion | Power Traces of AES Encryption | SVMs, DTs, RFs | Power Traces from DPA Contest Website and TeSCASE Website | Levina, Sleptsova, Zaitsev (2016) |
| Evasion; Oracle | Blackbox | Digital Watermarking; Break Our Watermarking System (BOWS) | DNN | Raise Image; Dresden Image | Quiring, Arp, and Rieck (2018) |
| Filter-aware Adversarial ML | Evasion | VGGNet model | DNNs | German Traffic Sign Recognition Benchmarks (GTSRB) Dataset, | Khalid et al. (2018a, b); Khalid et al. (2019) |
| Mimicry | Evasion; Training and Testing Data Sets | Gradient Descent Component | SVM; Neural Networks | MNIST Handwritten Digits; Contagio Dataset | Biggio et al. (2017) |
| Mimicry | Evasion; Gradient-Based | Gradient Descent–Kernel Density Estimation (GD-KDE); PDFRate System for Detecting Malicious PDF Files | SVM; RF | Contagio; Surrogate; Attack | Srndic and Laskov (2014) |
| Tree Ensemble | Evasion | Mixed Integer Linear Program (MILP); Approximate Evasion Algorithm | Boosted Trees; RF | MNIST Handwritten Digits Dataset | Kantchelian, Tygar, & Joseph (2016) |
| Problem Space; Feature Space | Evasion; Training Data | EvadeML - ML-Based PDF Malware Detector; MalGAN | SL2013; HIDOST | Contagio Archive | Tong et al. (2018) |
| Model Extraction | Evasion; Blackbox | PDF Malware Classifiers; Robustness of Classifiers | GP; PDFRate (Random Forest); HIDOST (SVM) | Contagio Archive | Xu, Qi, and Evans (2016) |
| Feature Squeezing | Adversarial Example | Feature Squeezing: Color Depth and Spatial Smoothing. | DNNs | MNIST; ImageNet; CIFAR-10 | Xu, Evans, and Qi (2018) |
| Information Leakage | Exploratory; Training Data | A Model: Meta-Classifier | ANNs; SVMs; HMM; DT | Internet Traffic Classifier; Speech Recognition Software | Ateniese et a. (2015) |
| Adversarial Examples | Evasion; Blackbox; Training/Testing | Forward Derivatives / Adversarial Saliency Maps; JSMA; Image Classification | DNN | MNIST Dataset; Handwritten Set | Papernot et al. (2016) |
| Adversarial Examples | Black-box | Reservoir Sampling; Remote ML Classifiers; | DNNs, LR, SVMs, DTs, kNNs | MNIST | Papernot, McDaniel, and Goodfellow (2016) |
| Adversarial Examples | Evasion | Neural Networks: Sigmoid; ReLU; RBFI | NNs | MNIST | Alfaro (2018) |
| Feature Selection | Evasion | Adversary-Aware Feature Selection Model | SVM with RBF Kernel | PDF malware; Spam dataset | Zhang et al. (2016) |
| A Threat Model | Categorization of Threats and Defenses | A Threat Model for categorizing attacks and defenses | Any | Any | Papernot et al. (2018; 2016b) |
| Adversarial Examples | Adversarial Examples | Prediction Errors; L-BFGS | Neural Networks | MNIST; ImageNet; YouTube Image Samples | Szegedy et al. (2014) |
| Attribute Inference | Evasion | AttriGard | LR, RF, Neural Networks | Google Play Apps | Jia and Gong (2018) |
| Reverse Engineering (RE) | Black-box | Metamodel Input (T-SNE); Adversarial Image Perturbations | NNs | MNIST; ImageNet Classifiers | Oh et al. (2018) |
| Availability | Causative; Training Data | Dictionary Attack; Focused Attack | SpamBayes | TREC corpus | Barreno (2008) |
| Integrity | Exploratory; Training Data | Learner Comparison with GMM, PMM, SVM | LDA | Enron email corpus; Email traces | Barreno (2008) |
| Hyperparameter Stealing | Exploratory | Theoretical and Empirical Evaluations, | RR, LR; SVM; NN | Diabetes, GeoOrig; Iris; Mandelon, etc. | Wang and Gong (2018) |
| Zeroth-Order Optimization | Black-box | Only Access to the Input (Images) and the Output (confidence scores) of a targeted DNN. | DNNs | MNIST; CIFAR | Chen et al. (2017b) |
| Targeted Noise Injection; Small Community | Black-box; Adversarial ML | Graph Clustering; Domain Name Generation Algorithms (DGAs) | Singular Value Decomposition (SVD) | US Telecommunication; a US University; a Threat Feed | Chen et al., 2017 |
| Adversarial Example | Adversarial Example - Audio | Speech Recognition Model | DNNs | Audio Phrases | Yakura and Sakuma (2018) |
| Adversarial Examples | Gradient-Based Attacks | Robustness of DNNs classifier; DeepFool Algorithm | Convolutional DNN | MNIST; CIFAR-10; ILSVRC 2012 | Moosavi-Dezfooli, Fawzi, & Frossard (2016) |
| Evasion | Gradient-based Attacks; Black-box | Android Malware Detection System | SVM; SVM-RBF; RFs | Drebin Data - Benign and Malicious Samples. | Melis et al. (2018) |
| Evasion | Fast Gradient Sign Method (FGSM) | FGSM; Fast-Jacobian Saliency Map Apriori (FJSMA); web-based application | CNN | MNIST | Norton and Qi (2017) |
| Poisoning | Direct Gradient Method | Generative Adversarial Network (GAN) | NNs; DNN | MNIST; CIFAR | Yang et al. (2017) |
| Hardware Trojan | A Framework | Threat Models and Taxonomy; General Framework | NNs; CNNs | MNIST; CIFAR | Clements and Lao (2018) |
| Evasion | Score-Based Attacks; Black-box | Black-box Variants of JSMA; One-Pixel; Greedy Local-Search | CNNs | MNIST; CIFAR; SVHN; STL | Narodytska and Kasiviswanathan (2016) |
| Adversarial Example | Black-box; Transfer-based | Attack a DNN by MetaMind (an API); API by Google, Amazon | DNNs | MNIST; Handcrafted; GTSRB | Papernot et al. (2017) |
| Adversarial Example | Black-box; Transfer-based | Ensemble-based Approaches | DNNs; ResNet; VGG; GoogLeNet | ILSVRC 2012 | Liu et al. (2017) |
| Universal Adversarial Perturbations | Score-Based Attacks | Generator Networks That Predict Adversarial; Universal Adversarial Networks (UANs) | NNs | CIFAR; ImageNet | Hayes & Danezis (2017) |
| Evasion | White-box; Adversarial Defense | Data Transformations | PCA; SVMs; DNNs; CNNs | MNIST image dataset; UCI Human Activity Recognition (HAR) dataset | Bhagoji et al. (2017a, b) |
| Optimization-based Methods | Query-Limited; Black-box | Bayesian Optimization Based Method | SVM; ANNs | Email Spam Dataset | Suya et al. (2017) |
| Evasion | Gradient-based | The Vulnerability of Malware Detection Methods - Raw bytes input | NNs | Malware Samples - VirusShare, Citadel and APT1. | Kolosnjaji et al. (2018) |
| Fast Gradient Sign Methods (FGSM) | Adversarial Example | Linear Perturbation | DNNs | MNIST | Goodfellow, Shlens, and Szegedy (2015); Ozdag (2018) |
| Poisoning (Stingray); Evasion | A General Framework | The FAIL Attacker Model | NNs | CIFAR; Drebin; Twitter vulnerability databases | Suciu et al. (2018) |
| Carlini-Wagner (CW) | Score-Based Attacks | Defeat Defensive Distillation; Deep fool, Fast Gradient Sign, and Iterative Gradient Sign. | NNs | MNIST; CIFAR; ImageNet | Carlini and Wagner (2017); Chen et al. (2017); Ozdag (2018) |
| Adversarial Example | Black-box | Houdini and CTC for Speech recognition, pose estimation and semantic segmentation. | Speech system | SSIM; Perceptibility | Cisse et al. (2017) |
| Adversarial Example | Human Perceptual System | Adversarial Example Attack Crafting<br>Method with Human Perceptual System | DNNs | MNIST; CIFAR-10 | Luo et al. (2018) |
| Adversarial Example - 3D | Physical-World Inputs | Expectation Over Transformation (EOT); InceptionV3 Classifier | NNs | 3D-Printing Models | Athalye et al. (2018) |
| Evasion | Gradient Descent (GD) | Gradient Descent | non-liner SVMs | MNIST; USPS; SpamBase | Han and Rubinstein (2017) |
| Projected Gradient Descent (PGD) | Adversarial Example | Robust Optimization | NNs | MNIST; CIFAR-10 | Madry et al. (2017); Ozdag (2018) |
| Basic Iterative Method (BIM) | Adversarial Example | Iterative FGSM; Label Leaking | DT, RF | ImageNet | Kurakin, Goodfellow, & Bengio (2017); Ozdag (2018) |
| Model Inversion | ML-as-a-service APIs. | ML-as-a-Service (MLaaS) API; Face Recognition | DT, NNs | FiveThirtyEight Survey; General Social Survey (GSS) | Fredrikson, Jha, Ristenpart (2015); Kumar & Mehta (2018) |
| Model Extraction | Blackbox; Path Finding | ML-as-a-Service (MLaaS) API | LR; DNN; DT | German Credit Data; Circles, Iris, Adult Data | Tramer et al. (2016); Wang and Gong (2018); Nguyen (2018) |
| Model Extraction | PLM-Based | Primitive Learning Modules (PLMs); RNN Model Evaluation; LSTM | RNN; LR; RF | Synthetic and Real-World Datasets | Xue and Chuah (2018) |
| Membership Inference | Model Extraction; Defenses | ML-as-a-Service (MLaaS) - Google Cloud Prediction API | MLP; LR; RF | MNIST, CIFAR-10, CIFAR-100, etc. | Salem et al. (2018) |
| Membership Inference; OR Attribute Inference | Information from Training Data | Connection to Overfitting and influence - Privacy of Training Data | LR, DT, CNNs Models | MNIST; CIFAR | Yeom et al. (2018) |
| Membership Inference | Black-box | Quantitative Assessment; ML-as-a-service | NNs; CNNs | MNIST; CIFAR; Texas Hospital - Stay Dataset, Adult Dataset | Shokri et al. (2017) |
| Backdoor | Primitive Learning Modules (PLM) -Based | Inception Model v3 | DNN | ImageNet; ISIC | Ji et al. (2018) |
| Adversarial Example | GAN - VAE | GAN Discriminator - Learned Feature Representations | VAE/GAN Models | Smaller Images; Enc, Dec and Dis | Larsen et al. (2016) |
| Adversarial Example | GAN - VAE | GAN discriminator | Three attacks; ReLU activation function; | MNIST, SVHN and CelebA | Kos, J., Fischer, and Song (2017) |
| Adversarial Example | GAN | The Generative Nets | Adversarial Nets; Semi-supervised learning | MNIST; TFD; CIFAR-10 | Goodfellow et al. (2014) |
| Adversarial Example | Single Perturbation | Adversarial Reprogramming | Models trained on ImageNet | MNIST; CIFAR-10 | Elsayed, Goodfellow, and Sohl-Dickstein (2018) |
| Adversarial Example | No - Security Through Obscurity | Locally Linear Embedding (LLE) - Non-Parametric Dimensionality Reduction | DNNs | MNIST; IMDB; Malware | Wang et al. (2017) |
| Randomized Prediction Games | Game Theory; Non-Cooperative | Randomized Prediction Games | SVM | Handwritten Digit Recognition; Trec2007; Malware in PDF Files. | Bulò et al. (2017) |
| Mimicry | Attack Models; Training Data | Gradient Descent and Kernel Density Estimation (GD- KDE) | SVMs, Bayesian, DT, RF | Sina Weibo | Kantarcioglu and Xi (2017) |
| Polymorphic Blending | Mimicry Attack | Byte-Frequency Based IDS; PAYL; N-gram | IDS PAYL | Network Traffic | Fogla et al. (2006) |
| Stackelberg Game - Nested | Free-Range; Targeted; Game Theory | A Single-step Two Players' Game; Adversarial Classifier Reverse Engineering (ACRE) | NB; SVM; RVM; HME | Text classification; PDF document; Feature Extraction | Kantarcioglu and Xi (2016; 2017); Zhou and Kantarcioglu (2016); Nguyen (2018) |
| A Game - Classifier and Adversary | Game Theory; Decision-Based | Adversarial Classifier System | NB; Adversarial-Classifiers | Ling-Spam; Email Spam Data | Dalvi et al. (2004) |
| A Game-Theoretic Framework | Game Theory | Distributed -SVM; Alternating Direction Method of Multipliers (ADMoM) | SVM | Rand; Spam; MNIST | Zhang and Zhu (2018) |
| A Single-Shot Prediction Game | Game Theory | Equilibria Prediction Models. | SVM; LR; SVM w/ Invariances | Email Spam Filter | Bruckner and Scheffer (2009) |
| Stackelberg Game | Game Theory | Stackelberg Prediction Game | SVM; LR | Email Spam Filter: ESP; Mailing List; TREC 2017, etc. | Bruckner and Scheffer (2011); Nguyen (2018) |
| Adversarial Classifier Reverse Engineering (ACRE) | A Framework | Randomized Classification Schemes; | NB; SVM | Enron data; Spam Classification | Vorobeychik and Li (2014) |
| Cross-Substitution | Evasion; | Adversarial Classifier Algorithm; Adversarial Models; Stackelberg game multi-adversary model (SMA) | NB; SVM; NNs | Enron data; Ling-Spam Spam; UCI data | Li & Vorobeychik (2014) |
| Good Words | Decision-Based Attacks | Adversarial Classifier Reverse Engineering (ACRE); Continuous and Boolean features | Linear Classifiers | Spam Filter Data; Dictionary Words | Lowd, D., & Meek., C. (2005 |
| Blind Newton Sensitivity | Sensitivity-Based | Non-Linear Optimization; Watermarking Schemes | ML Detectors | Synthetic Images | Comesaña, Pérez-Freire, & Pérez-González (2006); Quiring, Arp, and Rieck (2018) |
| Poisoning; Greedy Optimal | Adversarial Noise | Online Centroid Anomaly Detection | Learning model: K-gram length; RBF kernel | Network Traffic | Kloft and Laskov (2012 |
| Poisoning - Boiling Frog | Adversarial Defenses; Network IDS | Statistical Machine Learning (SML); Antidote; PCA-subspace | PCA | Network Traffic | Rubinstein et al. (2009) |

## Appendix B

Table 1
A list of Abbreviations

| Abbreviation | Full-form |
|---|---|
| AE/P | Adversarial Examples/Perturbations |
| API | Application Programming Interface |
| CART | Classification and Regression Tree |
| CD | Coordinate Descent |
| CNN | Convoluted Neural Networks |
| CNTK | Cognitive Toolkit (Microsoft) |
| CPS | Cyber–physical System |
| CV | Computer Vision |
| DNN | Deep Neural Networks |
| DoS | Denial of Service |
| DR | Dimensionality Reduction |
| DL | Deep Learning |
| DT | Decision Tree |
| FNN | Feedforward Neural Network |
| GAN | Generative Adversarial Network |
| GD | Gradient Descent |
| GMM | Gaussian Mixture Model |
| GP | Genetic Programming |
| HMM | Hidden Markov Model |
| ICS | Industrial Control System |
| IDS | Intrusion Detection System |
| IoT | Internet of Things |
| JSMA | Jacobian Saliency Map Approach |
| KNN | K-Nearest Neighbor |
| LDA | Latent Dirichlet Allocation |
| LR | Linear/Logistic Regression |
| ML | Machine Learning |
| MLP | Multi-Layer Perceptron |
| N/A | Not Applicable |
| NB | Naive Bayes |
| NN | Neural Network |
| NTLK | Natural Language Toolkit |
| PCA | Principal Component Analysis |
| PGA | Projected Gradient Ascent |
| PGD | Projected Gradient Descent |
| RBF | Radial Basis Function kernel |
| RF | Random Forest |
| RR | Ridge Regression |
| RVM | Relevance Vector Machine |
| SML | Statistical Machine Learning |
| SVD | Singular Value Decomposition |
| SVM | Support Vector Machine |
| VAE | Variational Autoencoder |